\documentclass[11pt]{article}

\usepackage[final]{acl}

\usepackage{times}
\usepackage{latexsym}

\usepackage[T1]{fontenc}

\usepackage[utf8]{inputenc}

\usepackage{microtype}

\usepackage{inconsolata}

\usepackage{graphicx}
\usepackage{todonotes}

\usepackage{booktabs, tabularx}
\usepackage{listings}
\usepackage{xcolor}
\usepackage{amsmath}
\usepackage{tabularx}
\usepackage[export]{adjustbox}
\usepackage{ragged2e}
\usepackage{enumitem}

\title{\includegraphics[height=5em,valign=c]{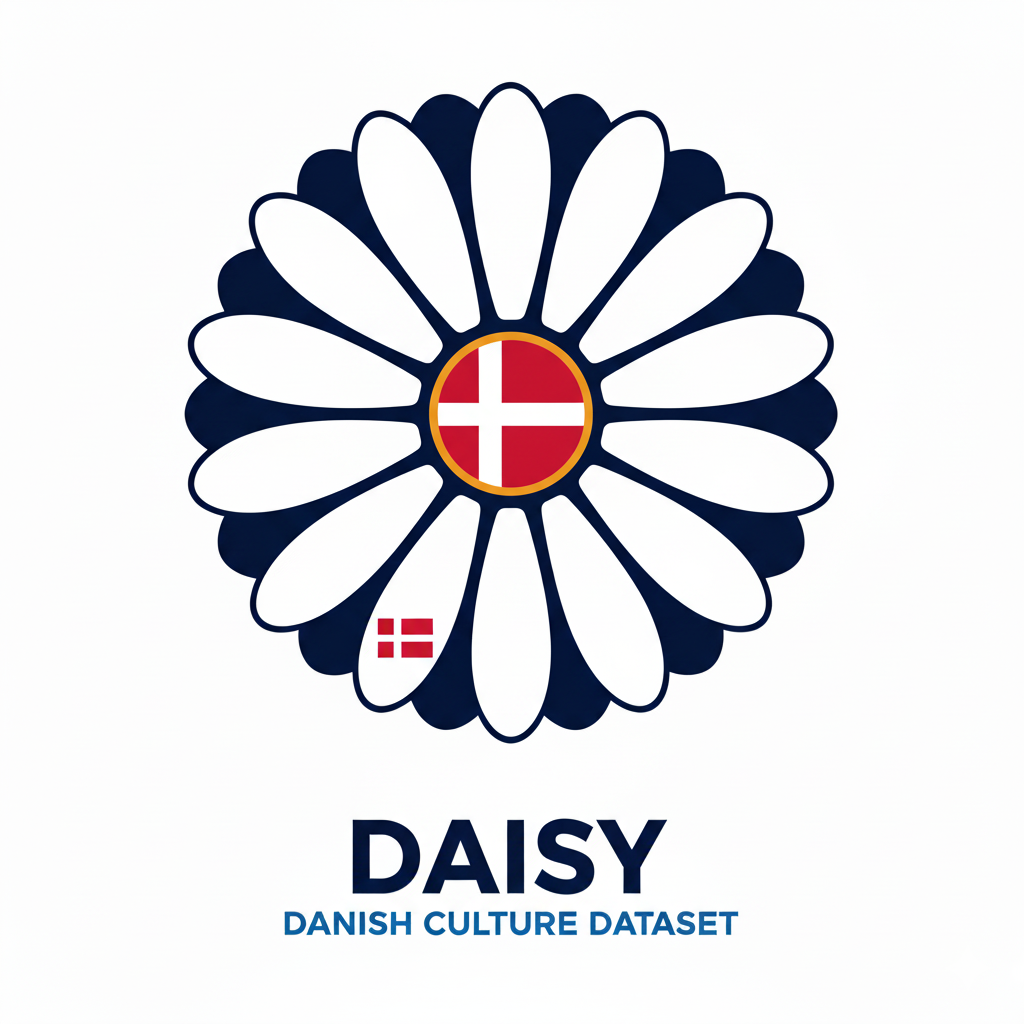}Daisy: A Benchmark for Danish Culture}

\author{Jacob Nielsen$^{*}$, Stine Lyngsø Beltoft$^{*}$, Peter Schneider-Kamp, Lukas Galke Poech \\
        University of Southern Denmark\\
        \texttt{\{jacn,stinelb,petersk,galke\}@imada.sdu.dk}}

\begin{document}
\maketitle
\renewcommand{\thefootnote}{\fnsymbol{footnote}}
\footnotetext[1]{Equal contribution.}
\renewcommand{\thefootnote}{\arabic{footnote}}
\begin{abstract}
We introduce Daisy, a factual knowledge benchmark for Danish cultural heritage, based on curated topics from the Danish Culture Canon 2006. For each artifact in the culture canon, we query the corresponding Wikipedia page and have a language model generate a diverse set of questions. Within each artifact, we sample both central and peripheral questions, testing not only mainstream information but also the deeper, defining elements of Danish cultural heritage as identified by the Canon committee. Each question–answer pair is manually approved or corrected, yielding a final dataset of 741 closed-ended question–answer pairs ranging from archaeological findings dated to 1300 BCE and 18th-century poems and musical pieces through to contemporary pop music, Danish design, and architecture. Baseline results on our benchmark show that contemporary language models (GPT-OSS-120B \& 20B, Llama-3.3-70B, Gemma3-27B, and Mistral-3.1-24B) perform surprisingly poorly on this task, despite the tested factual knowledge being available on Wikipedia. The best-performing model, Llama-3.3-70B, attains only 0.17 BLEU and an F1 score of 0.27, demonstrating the benchmark's difficulty. 
The questions, model benchmarks, and the evaluation tool are publicly available at 
\href{https://github.com/schneiderkamplab/SDU-Daisy}{github.com/schneiderkamplab/SDU-Daisy}
and under \href{https://huggingface.co/datasets/schneiderkamplab/SDU-Daisy}{schneiderkamplab/SDU-Daisy} on Hugging Face Datasets.
\end{abstract}

\section{The Missing Danish Benchmarks}\label{sec1}
Large Language Models (LLMs) have demonstrated remarkable capabilities across diverse natural language processing tasks~\cite{bubeck2023sparks}, yet their cultural knowledge has been so far understudied, particularly for non-English linguistic contexts~\cite{tao2024cultural}. While considerable research has examined the performance of LLMs on English-language benchmarks \cite{phan2025humanity} and culturally Anglo-centric tasks \cite{hershcovich-etal-2022-challenges}, the representation and understanding of non-dominant cultures in these models has, until recently, received comparatively little attention~\cite{liu2025culturally}. This disparity is especially pronounced for low-resource languages, where limited training data and evaluation frameworks compound the challenges of assessing and improving model performance. 
Danish exemplifies these challenges. Despite being spoken by approximately six million people and representing a rich cultural tradition spanning millennia, Danish is considered a low- to medium-resource language in contemporary AI research. Crucially, no standardized benchmark exists for assessing Danish cultural heritage in LLMs. This absence creates substantial obstacles for both model evaluation and development: researchers and practitioners lack systematic methods to measure how well models capture Danish cultural knowledge, historical context, and linguistic nuances that extend beyond mere language proficiency. Without such benchmarks, the development of Danish language models proceeds without adequate feedback mechanisms, resulting in minimal focus on preserving and accurately representing Danish cultural heritage in AI systems.

\begin{figure*}[ht!]
    \centering
    \includegraphics[width=\linewidth]{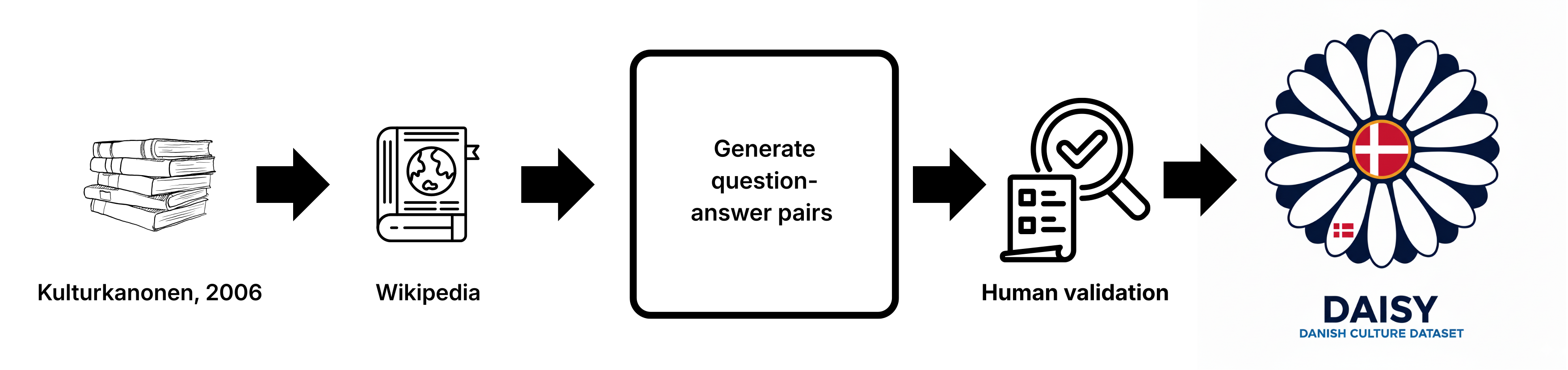}
    \caption{Overview of the stages in Daisy's Generation Pipeline.}
    \label{fig:Daisy-pipeline}    
\end{figure*}

The need for culturally-grounded evaluation frameworks extends beyond academic interest. As LLMs become increasingly integrated into educational systems, cultural institutions, and public services as well as different types of analysis tasks, their ability to accurately represent and engage with local cultural knowledge becomes essential. Misrepresentation or omission of cultural content can perpetuate biases, erode cultural identity, and limit the utility of these technologies for non-English speaking populations. For smaller linguistic communities, the risk is particularly acute: without dedicated evaluation frameworks, model developers may overlook culturally significant content in favor of more widely represented cultural materials.

Here, we introduce Daisy\footnote{Daisy is the nickname of Queen Margrethe II of Denmark and the name of the Danish national flower.}.
To create Daisy, we leverage the Kulturkanon, a collection of Danish cultural heritage (details in \S\ref{sec2}), as a foundational reference to introduce the first culturally-grounded benchmark for evaluating Danish language models (see Fig.~\ref{fig:Daisy-pipeline}). By grounding our evaluation framework in an authoritative and widely-recognized cultural reference, we address a critical gap in multilingual and multicultural AI assessment~\cite{liu2025culturally}. This benchmark enables assessment of how well LLMs capture Danish cultural knowledge, providing researchers and practitioners with essential tools for model development, comparison, and improvement. In sum, our main contributions are:

\begin{itemize}[noitemsep]
    \item A new benchmark  testing factual knowledge of Danish cultural heritage based on authoritative materials
    \item Baseline evaluations of contemporary LLMs on this benchmark
    \item Insights into the challenges and opportunities for improving cultural representation in low-resource language modeling
\end{itemize}

We emphasize that Daisy targets factual cultural knowledge -- recall of concrete attributes of canonical works -- rather than broader cultural reasoning or competence.
In the following, we first briefly describe the Danish Culture Canon (\S\ref{sec2}), before we introduce the Daisy benchmark (\S\ref{sec:daisy}).
Subsequently, we provide baseline results on Daisy (\S\ref{sec:results}) and enter a discussion to contextualize Daisy with related previous resources (\S\ref{sec:discussion}), before we conclude.

\section{The Danish Culture Canon - Kulturkanon}\label{sec2}
The Kulturkanon (Culture Canon) \cite{kulturkanon2006} was established in 2006 by the Danish government, the Cultural Ministry, following a national initiative that began in 2005. This canonical collection represents an authoritative, curated compilation of Danish cultural heritage designed to define and preserve the nation's cultural identity for future generations. The canon encompasses 108 works distributed across eight cultural domains: architecture, visual arts, design and crafts, film, literature, music (popular and sheet music), and theater. The temporal scope of the Kulturkanon is remarkably broad, featuring artifacts ranging from archaeological findings dated to 1300 BCE through contemporary works, thereby providing a comprehensive historical perspective on Danish cultural heritage.

The selection process for the Kulturkanon involved extensive expert consultation across multiple cultural sectors, with committees of distinguished scholars, practitioners, and cultural authorities deliberating on representative works within each domain. This rigorous curation process, combined with official governmental endorsement, establishes the Kulturkanon as the definitive reference for Danish cultural heritage. The canon's explicit pedagogical purpose, to be integrated into educational curricula and serve as a foundation for cultural literacy, further reinforces its authoritative status within Danish society. 

The Kulturkanon's suitability as a benchmark foundation stems from several key characteristics. First, its official status and broad expert consensus provide legitimacy and representativeness that user-generated or algorithmically-derived datasets cannot match. Second, the canon's explicit design as a reference for cultural knowledge aligns directly with the evaluation objectives of measuring LLM understanding of Danish culture. Third, the diversity of domains and historical periods represented ensures comprehensive coverage of Danish cultural heritage rather than narrow topical focus. Finally, the canon's established role in Danish education and cultural discourse means that knowledge of these works represents genuine cultural competence rather than obscure trivia, making it an ecologically valid basis for evaluation.

Kulturkanon primarily corresponds to the ideational elements of \citet{liu2025culturally}'s taxonomy, especially the subcategory of knowledge/artifacts, since it is essentially a curated collection of canonical works. However, Kulturkanon also implicitly involves certain social elements of culture: by selecting and elevating particular works as nationally significant, it participates in shaping collective memory, national identity, and shared cultural reference points. In this way, Kulturkanon contributes to the social meaning and institutional framing of culture, even though it does not explicitly catalogue everyday practices, linguistic norms, communicative conventions, or behavioral patterns. Thus, while the canon is firmly rooted in ideational content, it also reflects social identity formation and cultural legitimation processes, if only indirectly.

\section{The Daisy Benchmark}\label{sec:daisy}
We build upon the Danish culture canon as an established resource, to create the Daisy benchmark. By creating questions and answers based on the Culture Canon we construct a dataset that represents the Danish cultural heritage as defined by the Danish state. This dataset allows us to assess how well language models -- particularly, large international ones: GPT-OSS-120B \& 20B~\citep{openai2025gptoss120bgptoss20bmodel}, Llama-3.3-70B~\citep{grattafiori2024llama3herdmodels}, Gemma3-27B~\citep{gemma_2025}, and Mistral-3.1-24B~\cite{mistralsmall31} -- perform when confronted with Danish culture-specific knowledge.

\subsection{Constructing the Dataset}
We generate the dataset by extracting the Wikipedia page for each work in the Culture Canon, which contains detailed information about each work in different lengths depending on the type of work. Question--answer pairs are generated by Gemma3 27B in a 4-bit quantization. The model is prompted with the respective page's content as input, asked to generate 5 random questions within a framework defined by our generation prompt, which we report in Appendix \ref{sec:appendic:gen_prompt_template}. This is essentially a sampling strategy within each work, generating a mix of central and peripheral questions for each work. We believe this makes the dataset a harder benchmark, not only testing mainstream information but in-depth knowledge about corner-stones defining the heritage of Danish culture. 

\begin{table*}[htb]
    \centering
    \small
    \caption{Sample questions with model answers. Correct answers are marked in \textbf{bold}. Cases in which the model provided no final answer are marked with `-'.}
    \label{tab:qa:example}
    \begin{tabularx}{\textwidth}{>{\raggedright\arraybackslash}X *{5}{>{\centering\arraybackslash}X}}
        \toprule
        \multicolumn{1}{c}{} 
            & \multicolumn{2}{c}{GPT-OSS} 
            & \multicolumn{1}{c}{Mistral-Small 3.1 Instruct 2503} 
            & \multicolumn{1}{c}{Llama 3.3}
            & \multicolumn{1}{c}{Gemma 3} \\
        \cmidrule(lr){2-3}
        \cmidrule(lr){4-6}
        Question 
            & 20B 
            & 120B 
            & 24B 
            & 70B 
            & 27B \\
        \midrule
        Hvem skrev salmen *De levendes Land*? 
            & Lars M. Jensen
            & Niels Bjerre
            & \textbf{N.F.S. Grundtvig}
            & \textbf{N.F.S. Grundtvig}
            & Thomas Kingo \\
        \midrule
        Hvem er sangskriveren bag *Kald det kærlighed*?
            & -
            & Rasmus Seebach
            & Lasse Lindorff
            & Rasmus Seebach
            & Rasmus Seebach \\
          \midrule
        Hvem skrev Telegrafgalop? 
            & Svend Gade
            & Tom Kristensen
            & B.S. Ingemann
            & \textbf{Hans Christian Lumbye}
            & Carl Nielsen\\
        \midrule
         Hvem er forfatteren til digtet Myggesang?
            & H.C. Andersen
            & Halfdan Rasmussen
            & Johannes Ewald
            & Johannes Ewald
            & Holger Danske. \\

        \bottomrule
    \end{tabularx}
\end{table*}

The pairs were validated by a single human annotator, a native Danish speaker with a background in anthropology. Gemma3 27B generated 741 candidate pairs across the 108 canon entities. Six entities ($5\%$) were omitted because of missing webpages and one due to incompleteness. We still get 741 pairs because we included individual works grouped under subtitles, e.g.: '4 revynumre' (i.e. four revue numbers) counts as one entity, but comprises 4 independent works.
During validation, each pair was reviewed for validity, factual correctness, and standalone interpretability (no implicit context).

\subsection{Evaluation Measures}
We evaluate the models by prompting them with instructions on the simple answering format. Broadly, the instructions specify that heights are given in meters, weights in kilograms, and dimensions (e.g., of a picture frame) as width by height. The models are instructed to only give the answer itself. We provide the prompt-template used for evaluation in Appendix \ref{sec:appendic:eval_prompt_template}. We evaluate the models by normalizing the answer  via case normalization and removing punctuation, articles, and extra white spaces. 
The F1 score is computed at the word level, formulated as:
\begin{equation}
    \text{Precision} = \frac{|\mathrm{tokens}_\mathrm{pred} \cap \mathrm{tokens}_\mathrm{gold}|}{|\mathrm{tokens}_\mathrm{pred}|}
\end{equation}

\begin{equation}
    \text{Recall} = \frac{|\mathrm{tokens}_\mathrm{pred} \cap \mathrm{tokens}_\mathrm{gold}|}{|\mathrm{tokens}_\mathrm{gold}|}
\end{equation}

\begin{equation}
    \text{F1} = \frac{2 \cdot \text{Precision} \cdot \text{Recall}}{\text{Precision} + \text{Recall}}
\end{equation}
We formulate it over token sets in order to be tolerant to small deviations in generation, even though the questions are closed-ended. By inspecting the outputs of each model, we can confirm that this strategy is sufficient.
The BLEU score is calculated with NLTK's \texttt{sentence\_bleu} function\footnote{https://www.nltk.org/\_modules/nltk/translate/bleu\_score.html}, given the predictions, ground truth and \texttt{SmoothingFunction().method4}$^{2}$ to account for the short lengths of generated texts, making the results better comparable. 
Despite limitations in special cases, we found using soft metrics BLEU and F1 to be preferable in pre-experiments, as these metrics score minor variations appropriately, e.g., variation in the use of apostrophes. 

\subsection{Characteristics of the Benchmark}
The benchmark consists of closed-ended QA pairs with a single predefined correct answer, spanning architecture, art, design \& crafts, movies, literature, music, popular music and performing arts. The nature of the questions is quite diverse and can be factual questions about an invention or art piece like its physical dimensions or where it is geographically from. There are also pairs containing questions on the author of a poem, song, or performing art -- and even the number of notes in musical pieces. Furthermore, there are questions about architecture, like buildings and bridges, including when they were built and opened. We provide examples in Table \ref{tab:qa:example}'s first column.

\section{Baseline Results}\label{sec:results}

\begin{table*}[ht!]
    \centering
    \small
     \caption{Baseline evaluation results on Daisy using prompt and dataset version 1.}
    \begin{tabular}{lcc}
    \toprule
    \textbf{Model} & \textbf{BLEU score} & \textbf{F1 score} \\
    \midrule
    openai/gpt-oss-20b & 0.062 & 0.112 \\
    openai/gpt-oss-120b & 0.126 & 0.211  \\
    google/gemma-3-27b-it & 0.123 & 0.193   \\
    meta-llama/Llama-3.3-70B-Instruct & \textbf{0.166} & \textbf{0.268}   \\
    mistralai/Mistral-Small-3.1-24B-Instruct-2503 & 0.124 & 0.202  \\
    \hline
    \end{tabular}
    \label{tab:qa:results}
\end{table*}

We report our baseline evaluations in Table \ref{tab:qa:results} and some examples in Table \ref{tab:qa:example}. None of the evaluated models performs very well, even though they are considered decent at generating correct Danish. Llama 3.3 70B, with a BLEU score of 0.166, outperforms all the newer models by a large margin. The second best scores are attained by GPT-OSS 120B with a BLEU score of 0.126. Gemma3 27B is often considered one of the better models in Danish, as measured on EuroEval\footnote{https://euroeval.com} \citep[formerly ScandEval;][]{smart2023scandeval}. However, also Gemma3 27B seems to lack factual knowledge about Danish culture, underlining not only the performance gap of Danish as a low-resource language, but also the gap between language and culture, which could be an artifact of synthetic training data.
Table \ref{tab:qa:example} shows the outputs of each model across 4 questions from Daisy. Generally, all models generate reasonable valid outputs, except for GPT-OSS-20B, which sometimes produces a reasoning trace exceeding 2,000 tokens before reaching an answer -- in which cases we drop the answer. 
Further results measured with an exact-match metric and including Danish foundation models are available in ~\citet{schneiderkamp2026dfmmimirv1open}.

\section{Discussion and Related Work}\label{sec:discussion}
Benchmarking cultural knowledge in language models has important implications for both society and the research community. First, having a measurement for the presence of Danish cultural heritage in existing models, can shed light on the missing cultural alignment and expected performance in existing models. Second, enabling benchmarking on curated datasets can guide researchers and practitioners in developing new models -- and enable foreign entities to evaluate their models. Last, it will support research in digital humanities in investigating the existence of Danish culture in models, as well as interactions with the models based on the model's measured cultural understanding. 

As reported in Table~\ref{tab:qa:results}, the models generally exhibit poor performance on Daisy, even though Danish Wikipedia was very likely included in their training data. We hypothesize that these shortcomings may stem from the limited representation of Danish-language and Denmark-specific content within the overall training data distribution, where such nationally bounded knowledge constitutes a relatively weak signal. In addition, preference alignment and safety tuning may further bias models toward cautious, generalized responses rather than confident enumeration of closed cultural lists. 

A natural concern is contamination: since questions derive from Wikipedia, and the evaluated models were almost certainly exposed to Danish Wikipedia during pre-training, high scores could reflect memorization rather than cultural knowledge. Notably, we observe the opposite. If contamination dominated, we would expect strong performance on precisely this material; instead, all models score poorly despite near-certain exposure. This dissociation is itself informative: it indicates that the relevant knowledge, even when present in the training data, is under-weighted, abstracted away, or not reliably retrieved for nationally bounded, low-frequency content. Daisy is therefore diagnostic not because the underlying facts are absent from training, but because surfacing them requires cultural knowledge to be represented and retrievable rather than merely seen. We acknowledge that fully disentangling memorization from retrieval would require held-out or post-cutoff cultural material, which we leave to future work.

We now discuss how Daisy relates to previously introduced language resources that were designed to evaluate language models on Danish culture.

\subsection{Danish Benchmarks}
While Danish NLP has seen notable development in recent years, existing benchmarks focus primarily on linguistic capabilities rather than cultural knowledge. ScandEval \cite{smart2023scandeval}, a natural language understanding benchmark, evaluates Danish models on tasks including named entity recognition, sentiment classification, linguistic acceptability, and question answering. The benchmark has been used to evaluate Danish language models like SnakModel \cite{zhang2025snakmodel} on both language and culture-specific tasks, though the cultural component remains limited. DaCy \cite{f975f4ce65944e3ea958578003cee622} provides a unified framework for Danish NLP with capabilities in named entity recognition, part-of-speech tagging, and dependency parsing, but does not explicitly assess cultural knowledge.
Recent work at the University of Copenhagen has developed a benchmark for evaluating the intrinsic reasoning capabilities of Danish LLMs based on semantic dictionaries \cite{pedersen-etal-2024-towards}, focusing on semantic reasoning rather than cultural content. The DaNLP repository~\cite{danlp2021} offers various Danish NLP resources and datasets, while multilingual evaluation frameworks like Okapi~\cite{lai2023okapi} include Danish among 26 languages for tasks like ARC \cite{allenai:arc}, HellaSwag \cite{zellers2019hellaswag}, and MMLU \cite{hendryckstest2021}, though these are translations of English benchmarks rather than culturally grounded assessments. To our knowledge, no existing Danish benchmark systematically evaluates cultural knowledge or heritage understanding, representing a significant gap that our work addresses.

DaKultur \cite{muller2025dakultur} represents the first cultural evaluation study for Danish, comprising $1{,}038$ interactions from 63 demographically diverse participants. The study employed a crowdsourcing methodology, where participants compose prompts requiring Danish cultural knowledge, which are then answered by three different language models and subsequently rated by participants through accept/reject judgments.
However, the crowdsourcing nature of the prompt generation reveals significant conceptual ambiguities regarding what constitutes culturally relevant knowledge. Several examples illustrate this concern: (1) “Kender du Faser?” (Do you know Faser?), which appears to reference niche recent pop culture rather than shared cultural understanding; (2) “Hvad bestiller danskerne på en cafe?” (What do Danes order at a café?), which elicits stereotypical behavioral patterns rather than deeper cultural values or norms; (3) ``Må man plukke svampe i skoven?'' (Is one allowed to pick mushrooms in the forest?), which tests legal or practical knowledge rather than cultural competence; and (4) ``Planlæg en fed dag i København med udendørs aktiviteter'' (Plan a cool day in Copenhagen with outdoor activities), which functions more as a practical recommendation task than a cultural evaluation.
These examples illustrate a general difficulty in crowdsourced cultural evaluation. Questions such as ``Hvordan er danske familier?'' (What are Danish families like?) demand broad generalizations that risk reducing complex, heterogeneous social structures to reductive stereotypes. Any response that purports to characterize ``Danish families'' as a monolithic entity necessarily obscures intra-cultural variation across class, region, generation, and individual family dynamics. This raises a fundamental question: does cultural competence require the ability to produce such generalizations, or does it instead demand recognition of cultural diversity and resistance to essentialization? The dataset's reliance on participant-generated prompts, while democratizing the evaluation process, may thus inadvertently conflate various forms of knowledge -- factual, practical, stereotypical, and cultural -- without adequately distinguishing between them.

A related consideration is the frequent use of the indfødsretsprøven (the Danish citizenship test) as a benchmark for ``Danish cultural knowledge'' in LLM evaluation \cite{Kirilova2025Indfodsretsproven}. Methodologically, its usefulness is limited: the test and its answer key circulate widely online, making it highly probable that most foundation models have already been exposed to it during training and therefore perform well for reasons unrelated to genuine cultural understanding. More importantly, there is a conceptual issue. While indfødsretsprøven is a state-administered instrument, it is designed to evaluate civic compliance and basic knowledge relevant for naturalization, not to provide a culturally or symbolically meaningful account of Danishness. In this sense, it operationalizes a narrow, politically contingent definition of culture. The Kulturkanon, although also a state-initiated project and therefore not free from ideological framing, represents a different category of cultural artifact: a curated attempt to articulate Denmark’s artistic, historical, and aesthetic heritage. Using it as a benchmark does not resolve the normative question of who gets to define culture, but it shifts the evaluative focus from administrative facts to culturally embedded works and references, offering a more substantively grounded, though still contestable, basis for assessing models’ cultural competence.

Danish Dynaword \cite{enevoldsen2025dynaword} is not directly focused on Danish culture, but rather provides curated Danish pre-training data, potentially implicitly containing cultural artifacts. 

\subsection{International Benchmarks}
Our work draws inspiration from recent international efforts to develop culturally-grounded evaluation frameworks for LLMs. CulturalBench \cite{chiu2025culturalbench} introduced 1,696 human-written and verified questions covering 45 global regions across 17 topics, revealing that even frontier models achieve only 61.5\% accuracy on challenging cultural questions. 

BLEnD \cite{myung2025blendbenchmarkllmseveryday} provides a complementary perspective by focusing on everyday cultural knowledge across 16 countries and 13 languages, with 52,600 question--answer pairs covering mundane aspects of daily life often absent from online sources. This emphasis on culturally-specific common knowledge that may not appear in training data resonates with our goal of evaluating genuine cultural understanding. 
GlobalOpinionQA~\cite{durmus2024towards} provides a foundation for cultural evaluation and shows how model opinions lean towards certain cultures.

MUREL (Multicultural Resource for Evaluating Language Models)~\cite{namazifard2025isolating}  is a curated dataset of 85.2M tokens spanning 6 cultures derived from 69 source datasets, which includes 11M tokens encoding Danish culture. The dataset spans ideational, linguistic, and social dimensions, supporting analysis of culture-specific phenomena, but does not explicitly cover factual knowledge regarding cultural heritage.

MultiWikiQA~\cite{smart2025multiwikiqa} is a reading comprehension dataset covering 306 languages, based on Wikipedia articles. The dataset conducts a crowdsourced human evaluation of the fluency of the generated questions in 30 of the languages, and hence targets neither cultural engagement nor cultural knowledge, in contrast to Daisy's focus on factual cultural knowledge.

Domain-specific cultural heritage benchmarks have emerged for other contexts, such as evaluation frameworks for Chinese intangible cultural heritage using data from authoritative national sources \cite{zhao2025ich}, showing the value of anchoring benchmarks in official cultural repositories.

Our approach of leveraging the Kulturkanon similarly draws on authoritative national cultural definitions while adapting the methodology to the specific context of Danish heritage and the comprehensive scope of the canon's domains.
The present work distinguishes itself from these precedents by combining the authoritative grounding of nationally-defined cultural heritage with comprehensive domain coverage and the specific challenges of evaluating low-resource language models on culturally-nuanced content. Our benchmark uniquely addresses the intersection of official cultural heritage, comprehensive artistic and intellectual domains, and the evaluation needs of Danish language technology development. 

Even in Humanity’s Last Exam~\cite{phan2025humanity}, a benchmark explicitly framed as testing ``the forefront of human knowledge'', the center of gravity remains firmly in the natural sciences and mathematically structured domains. This reveals a persistent assumption in AI evaluation: that knowledge is best represented through problems with clear formal structure and objectively scorable answers. But knowledge is far more expansive than mathematics. If we ever reach a point where a “last exam” for humanity is meaningful, it will have to include the interpretive, historical, social, and cultural forms of understanding that shape human life. Humanities disciplines -- history, sociology, anthropology, literature, philosophy -- capture dimensions of meaning, value, and context that cannot be reduced to equations or symbolic manipulation. To take the idea of such an exam seriously, we must broaden our evaluative horizons to include precisely the kinds of cultural and interpretive competencies that define human experience, rather than only the domains where machines are already most comfortable.

\section{Conclusion}
We have introduced Daisy, a cultural closed-ended question--answer benchmark, rooted in the Danish Culture Canon spanning 741 question--answer pairs across eight cultural domains. We tested five contemporary models on Daisy that are popular or perform well on Danish language benchmarks -- all performing poorly despite being likely trained on data from Danish Wikipedia.
Daisy serves as an initial cultural beacon within the Danish Foundation Models initiative, illuminating the current state of development of Danish‑specific models, and, more generally, providing guidance for the wider community and future generations of models.
We plan to extend the fleet of Danish culture datasets and welcome contributions from sub-domain experts.

\section*{Limitations}

\paragraph{Scope.}
Daisy is centered around the Danish Culture Canon. Although a well-established resource, the Culture Canon is naturally limited in its scope. We do not argue that the Canon would fully contain or represent the whole Danish cultural heritage. Culture is far more diverse and continually evolving alongside society.
This also means that the culture canon, dated to 2006, does not contain newer cultural elements. However, this also means the artifacts within the benchmark have been digested and can be expected to be represented so generally that they should be recognizable by an LLM excelling in Danish and Danish culture. 

\paragraph{Representativeness.} Beyond temporal scope, the Kulturkanon is a 2006 state-curated artifact and thus a contestable, partial operationalization of ``Danish culture.'' It privileges institutionally sanctioned, ideational heritage over living, everyday, and subcultural practice, and reflects the value judgments of its selection committee. Daisy inherits these framing choices. We therefore position Daisy as measuring knowledge of a specific, authoritative cultural canon rather than Danish culture in its full breadth.

\paragraph{Size.} With 741 pairs, Daisy is modest in size and best understood as an initial, extensible resource rather than a large-scale benchmark; fine-grained statistical comparison between closely matched models should be treated cautiously. We commit to ongoing expansion, including the children's canon and additional domains.

\paragraph{Single-Prompt Evaluation.}
Evaluating generative models is hard. Our benchmarking depends on the prompt that gives the models instruction on the task as well as the expected output format. We report the exact template in Appendix \ref{sec:appendic:eval_prompt_template}. This prompt is not necessarily optimal for all models, nor necessarily optimal at all. Therefore, we report it as \emph{Prompt Template Version 1}. We plan to iterate on the prompt in future work and call for contributions to improve and expand the Daisy benchmark. 

\section*{Ethical Considerations}
The measurement of culture remains a contentious endeavor within the social sciences, with scholars rightfully cautioning against the reductionism inherent in quantifying such a multifaceted and dynamic construct \cite{merry2016seductions, clifford1986writing}. For LLMs to effectively understand and generate context-specific language, however, they must encode some representation of cultural knowledge, the shared meanings, values, and communicative norms that shape human expression across different communities. \citet{liu2025culturally} provide a comprehensive taxonomy of cultural elements in NLP, identifying key dimensions including cultural concepts, knowledge, norms, values, and artifacts that language models must navigate, as detailed in \S\ref{sec2}. While the critique that culture resists simple quantification is well-founded, the alternative -- avoiding systematic assessment altogether -- is untenable if we aspire to develop AI systems that authentically reflect human diversity rather than defaulting to narrow cultural frames.
Culture is not static but continuously evolving, shaped by historical forces, technological change, and intercultural exchange. This dynamic mirrors the rapid development cycle of AI systems themselves, suggesting a productive parallel: can we develop LLMs that evolve alongside the cultures they serve? Addressing this question requires an initial framework, one that begins with established dimensions of national culture \cite{hofstede2011dimensionalizing} not because these dimensions capture the totality of cultural experience, but because they provide empirically grounded starting points for moving beyond the WEIRD (Western, Educated, Industrialized, Rich, Democratic) paradigm \cite{henrich2010weirdest} that dominates current AI development. The major commercial LLMs, including ChatGPT and similar systems, demonstrably reflect American or broadly Western cultural assumptions and ideological orientations \cite{tao2024cultural, cao2023assessing, naous2024having}. To develop more culturally contextualized systems, we must first establish methods to systematically measure and compare cultural representations, a pragmatic necessity given that AI systems require operationalizable metrics for training and evaluation.

\section*{Acknowledgements}
This research was supported by the Danish Foundation Models project, funded by the Ministry of Science, Higher Education and Digital Affairs.

\bibliography{custom}

\cleardoublepage
\appendix

\section{Generation Prompt Template}\label{sec:appendic:gen_prompt_template}
We use the generation prompt template, inducing the context from a given document, with a set of rules, which is formulated based on empirical testing.

\begin{quote}
\ttfamily\small
CUSTOM\_CLOSED\_QA\_PROMPT\_TEMPLATE = """TEXT: \{document\}
\# REGLER:
1. Baseret på teksten ovenfor, lav 5 spørgsmål med svar (question answer pairs).
2. Lav spørgsmålene så "diverse" and relavante til TEXT som muligt. Fokusér på
   det mest fundamentale vinden i TEXT. Spørgsmålene skal være med et definitivt
   svar. Angiv svaret direkte uden beskrivelse.
3. Du skal IKKE lave spørgsmål omkring kulturkanonen, forkusér på det mest
   fundamentale i teksten.
4. Husk at bruge anførselstegnene korrekt.
5. Brug personers fulde navne og brug ikke forkortelser. Eksempel: Grundtvig og
   n.f.s grundtvig skal skrives Nikolaj Frederik Severin Grundtvig
6. Svarene skal være korte og præcise, helst et enkelt ord eller en kort sætning.
7. Svarene må IKKE indeholde kommaer.
8. Svarene må IKKE indeholde flere svarmuligheder.
9. Svarene må IKKE indeholde beskrivelser eller yderligere information, kun det
   direkte svar.
10. Svarene må IKKE indeholde referencer til andre personer, steder eller
    begivenheder
11. Svarene må IKKE indeholde ja/nej svar.
12. Svarene må IKKE indeholde delvise svar eller antydninger, kun det fulde og
    korrekte svar.
13. Svarene må IKKE være indforstået eller kræve yderligere forklaring.

\#\# Allowed structure examples:
- "I hvilket år blev rytterstatuen af Frederik V rejst i Frederiksstaden?","1771"
- "Hvad er navnet på den lille landsby, der ligger nær Glorup, og som har en række
  boliger i cottage-stil?","Svindinge"

\#\# NOT ALLOWED structure examples:
- "Hvem har inspireret Asger Jorn til at lave maleriet Stalingrad?",
  "Pablo Picasso, samt Asger Jorns personlige overvejelser om historie og politik,
   og beskrivelser af Slaget om Stalingrad fra en af hans venner, Umberto Gambetta."
- "Hvem indviede vejforbindelsen over Storebælt i 1998?",
  "Dronning Margrethe indviede vejforbindelsen."
- "Er Mariebjerg Kirkegård medtaget i den danske kulturkanon?",
  "Ja, Mariebjerg Kirkegård er medtaget i den danske kulturkanon i kategorien
   arkitektur."

Output format (strictly, for parsing purposes):
Spørgsmål 1: [Insert question or task here]
Svar 1: [Insert answer or response here]

Spørgsmål 2: [Insert question or task here]
Svar 2: [Insert answer or response here]

Spørgsmål 3: [Insert question or task here]
Svar 3: [Insert answer or response here]

Spørgsmål 4: [Insert question or task here]
Svar 4: [Insert answer or response here]

Spørgsmål 5: [Insert question or task here]
Svar 5: [Insert answer or response here]
"""
\end{quote}

\section{Evaluation Prompt Template}\label{sec:appendic:eval_prompt_template}
We use the generation evaluation template, instructing the model to answer within a given rule-set. We do this instead of using a structured output but still get a form of normalization to base our evaluation on.
\begin{quote}
\ttfamily\small
PROMPT\_TEMPLATE = """
    Besvar spørgsmålet med kun det direkte svar, uden forklaring om hvorfor.
    Regelsæt:
    - Svar kun på dansk.
    - Hvis svaret er i højde, svar i meter (m).
    - Hvis svaret er i vægt, svar i kilogram (kg).
    - Hvis svaret er om en størrelse, svar i centimeter (cm).
      Fx Hvor stort er maleriet Mona Lisa? Svar: 77 cm x 53 cm.
    - Hvis svaret er en person angiv den måde de typisk bliver angivet på
      i danske tekster.

    \textbackslash n\textbackslash nSpørgsmål: \{question\}\textbackslash nSvar:"""
\end{quote}

\section{More Example Questions}\label{sec:appendic:examples}
We provide more examples in Table \ref{tab:qa:more_example}.
 \begin{table*}[htbp]
    \centering
    \small
    \caption{Sample questions with model answers. Correct answers are marked in \textbf{bold}.}
    \label{tab:qa:more_example}
    \begin{tabularx}{\textwidth}{>{\raggedright\arraybackslash}X *{5}{>{\Centering\arraybackslash}X}}
        \toprule
        \multicolumn{1}{c}{} 
            & \multicolumn{2}{c}{GPT-OSS} 
            & \multicolumn{1}{c}{Mistral-Small 3.1 Instruct 2503} 
            & \multicolumn{1}{c}{Llama 3.3}
            & \multicolumn{1}{c}{Gemma 3} \\
        \cmidrule(lr){2-3}
        \cmidrule(lr){4-6}
        Question 
            & 20B 
            & 120B 
            & 24B 
            & 70B 
            & 27B \\
        \midrule
        Hvornår blev Solvognen fundet?
            & 1907
            & \textbf{1902}
            & \textbf{1902}
            & \textbf{1902}
            & \textbf{1902} \\
        \midrule
        Hvem har komponeret musikken til 'Gå med i lunden'?
            & Børge Christensen
            & Traditionel (ingen kendt komponist)
            & Thomas Koppel
            & H.C. Lumbye
            & Ludvig Holstein-Ledreborg \\
        \midrule
        Hvem skrev Ditte Menneskebarn?
            & Jens Peter Jacobsen
            & Martin A. Hansen
            & Børge Houmann
            & \textbf{Martin Andersen Nexø}
            & \textbf{Martin Andersen Nexø} \\
        \midrule
        Hvem designede Pantonstolen?
            & Hanna Panton
            & \textbf{Verner Panton}
            & \textbf{Verner Panton}
            & \textbf{Verner Panton}
            & \textbf{Verner Panton} \\
        \midrule
        Hvad er Fingerplanen en plan for?
            & Fingerplanen er en plan for udviklingen af Finger-regionen i Danmark.
            & en plan for den overordnede by‑ og transportudvikling i hovedstadsregionen
            & Fingerplanen er en plan for landets arealudnyttelse
            & \textbf{Fingerplanen er en plan for udviklingen af det offentlige transportsystem i København.}
            & Hovedstadsregionens udvikling. \\
        \midrule
       Hvornår blev Vor Frue Kirke opført?
            & 1862
            & 1829 
            & 1182-1220.
            & Omkring 1100-1300-tallet
            & 1200-tallet. \\
        \midrule
        I hvilket år startede udgivelsen af Flora Danica?
            & 1791
            & \textbf{1761}
            & \textbf{1761}
            & \textbf{1761}
            & 1605 \\
        \midrule
        Hvem er kunstneren bag skulpturen 'Stående kvinde'?
            & Kunstneren er Søren Møller.
            & Jens Galschiøt
            & \textbf{Astrid Noack}
            & \textbf{Astrid Noack}
            & Barbara Hepworth \\      
        \bottomrule
    \end{tabularx}
\end{table*}
\end{document}